# A Sequential Algorithm for Training Text Classifiers

David D. Lewis (*lewis@research.att.com*) and William A. Gale (*gale@research.att.com*)

AT&T Bell Laboratories; Murray Hill, NJ 07974; USA

### Abstract

The ability to cheaply train text classifiers is critical to their use in information retrieval, content analysis, natural language processing, and other tasks involving data which is partly or fully textual. An algorithm for sequential sampling during machine learning of statistical classifiers was developed and tested on a newswire text categorization task. This method, which we call uncertainty sampling, reduced by as much as 500-fold the amount of training data that would have to be manually classified to achieve a given level of effectiveness.

## 1 Introduction

Text classification is the automated grouping of textual or partially textual entities. Document retrieval, categorization, routing, filtering, and clustering, as well as natural language processing tasks such as tagging, word sense disambiguation, and some aspects of understanding can be formulated as text classification. As the amount of online text increases, the demand for text classification to aid the analysis and management of text is increasing.

One advantage of formulating text processing tasks as classification is that methods from statistics and machine learning can be used to form text classifiers automatically. While using machine learning does require manually annotating training data with class labels, this annotation takes less skill and expense than, for instance, building classification rules by hand [1].

There is often more text available than can be economically labeled, so a subset or *sample* of the data must be chosen to label.[1] Random sampling [3] will usually not be effective. If only 1 in 1000 texts are class members (not atypical), and only 500 texts can be labeled, then a random sample will usually contain 500 negative examples and no positive ones. This will not support training a classifier to distinguish positive from negative examples.

Relevance feedback [4] does a kind of nonrandom sampling. In effect, users are asked to label those texts that the current classifier considers most likely to be class members. This approach, which we might call *relevance sampling*, is a reasonable strategy in a text retrieval context, where the user is more interested in seeing relevant texts than in the effectiveness of the final classifier produced. Relevance feedback has also been proposed for finding examples of unusual word senses [5]. However, relevance feedback has many problems as an approach to sampling. It works more poorly as the classifier improves, and is susceptible to selecting redundant examples.

Relevance sampling is a sequential approach to sampling, since the labeling of earlier examples influences the selection of later ones [6]. This paper describes an alternative sequential approach, *uncertainty sampling*, motivated by results in computational learning theory. Uncertainty sampling is an iterative process of manual labeling of examples, classifier fitting from those examples, and use of the classifier to select new examples whose class membership is unclear. We show how a probabilistic classifier can be trained using uncertainty sampling. A test of this method on a text categorization task showed reductions of up to 500-fold in the number of examples that must be labeled to produce a classifier with a given effectiveness.

## 2 Learning with Queries

A classifier can often be learned from fewer examples if the learning algorithm is allowed to create artificial examples or *membership queries* and ask a teacher to label them [7, 8].[2] In many learning tasks creation of artificial examples is no problem. However, an artificial text created by a learning algorithm is unlikely to be a legitimate natural language expression, and probably would be uninterpretable by a human teacher.

---

[1] An exception is when large quantities of previously labeled text are available, as when automated text categorization is being deployed to replace or aid an existing staff of manual indexers [2].

[2] In this paper "queries" always refers to membership queries, not text retrieval queries.

1. Create an initial classifier
2. While teacher is willing to label examples
   (a) Apply the current classifier to each unlabeled example
   (b) Find the $b$ examples for which the classifier is least certain of class membership
   (c) Have the teacher label the subsample of $b$ examples
   (d) Train a new classifier on all labeled examples

Figure 1. An algorithm for uncertainty sampling with a single classifier.

Recently, several algorithms for learning via queries have been proposed that filter existing examples rather than creating artificial ones [9, 10, 11]. These algorithms ask a teacher to label only those examples whose class membership is sufficiently "uncertain". Several definitions of uncertainty have been used, but all are based on estimating how likely a classifier trained on previously labeled data would be to produce the correct class label for a given unlabeled example. Looking at this as a sampling method rather than a querying one, we call this approach *uncertainty sampling*.

Seung, Opper, and Sompolinsky [11] present a theoretical analysis of "query by committee" (QBC), an algorithm that, for each unlabeled example, draws two classifiers randomly from the *version space*, i.e. the set of all classifiers consistent with the labeled training data [12]. An infinite stream of unlabeled data is assumed, from which QBC asks the teacher for class labels only on those examples for which the two chosen classifiers disagree.

Freund, Seung, Shamir, and Tishby extend the QBC result to a wide range of classifier forms [13]. They prove that, under certain assumptions, the number of queries made after examining $m$ random examples will be logarithmic in $m$, while generalization error will decrease almost as quickly as it would if queries were made on all examples. More precisely, generalization error decreases as $O(1/m)$. Therefore, in terms of the number of queries, generalization error decreases exponentially fast.

This is a provocative result, since it implies that the effect of training on labeled data can be gotten for the cost of obtaining unlabeled data, and labeling only a logarithmic fraction of it. However, the QBC assumptions include that the data is noise free, a perfect deterministic classifier exists, and that it is possible to draw classifiers randomly from the version space, all of which are problematic for real world tasks. The effectiveness of QBC and related methods on real world tasks remains to be determined.

A heuristic alternative to QBC's random drawing of classifiers from the version space is to let the learning algorithm do what it always does—pick a single classifier from the version space. If the classifier can not only make classification decisions, but estimate their certainty, the certainty estimate can be used to select examples.

A single classifier approach to uncertainty sampling has several theoretical failings, including underestimation of true uncertainty, and biases caused by nonrepresentative classifiers [9, 10]. On the other hand, experiments using a single classifier to make arbitrary queries [14] or select subsets of labeled data [8, 15] have shown substantial speedups in learning. Relevance sampling, which has proven quite effective for text retrieval, also uses a single classifier.

## 3 An Uncertainty Sampling Algorithm

Figure 3 presents an algorithm for uncertainty sampling from a finite set of examples using a single classifier. Ideally $b$, the number of examples selected on each iteration, would be 1, but larger values may be appropriate if scoring and selecting examples is expensive. This algorithm can be used with any type of classifier that both predicts a class and provides a measurement of how certain that prediction is. Probabilistic, fuzzy, nearest neighbor, and neural classifiers, along with many others, satisfy this criterion or can be easily modified to do so. Perhaps the most difficult requirement is that measurements of relative certainty be produced even when the classifier was formed from very few training examples.

Uncertainty sampling is similar to the strategy of training on misclassified instances [16, 17]. The difference is that when data is not labeled we must use the classifier itself to guess at which examples are being misclassified. Note that the initial classifier plays an important role, since without it there may be a long period of random sampling before examples of a low frequency class are stumbled upon.

## 4 A Probabilistic Text Classifier

In this section we describe a classifier form which produces estimates of $P(C_i|\mathbf{w})$, the posterior probability that an example with pattern $\mathbf{w}$ belongs to class $C_i$. Estimates of this probability can be used both to decide when an example should be assigned to a class, and to estimate how likely it is that the classification will be correct. We describe how the classifier is trained and how we use it for uncertainty sampling and classification.

## 4.1 A Probabilistic Classifier

Classifiers which estimate the posterior probability via Bayes' Rule:

$$P(C_i|\mathbf{w}) = \frac{P(\mathbf{w}|C_i) \times P(C_i)}{\sum_{j=1}^{q} P(\mathbf{w}|C_j) \times P(C_j)} \tag{1}$$

have been applied to a variety of text classification tasks, including text retrieval [18], text categorization [19, 20], and word sense identification [5]. Here the $C_i$ are a disjoint and exhaustive set of classes to which an example may belong, and $\mathbf{w} = (w_1, ..., w_d)$ is an observed pattern.[3] $P(\mathbf{w}|C_i)$ is the conditional probability that an example has pattern $\mathbf{w}$ given that it belongs to class $C_i$, while $P(C_i)$ is the prior probability that an example belongs to class $C_i$.

In this paper, we treat only the case $q = 2$, so there are two classes $C_1 = C$ and $C_2 = \bar{C}$, with $P(\bar{C}) = 1 - P(C)$. In this case it is useful to express the relative posterior probabilities of $C$ and $\bar{C}$ as an odds ratio:

$$\frac{P(C|\mathbf{w})}{P(\bar{C}|\mathbf{w})} = \frac{P(C)}{P(\bar{C})} \times \frac{P(\mathbf{w}|C)}{P(\mathbf{w}|\bar{C})} \tag{2}$$

Given the huge number of possible $\mathbf{w}$'s, estimation of $P(\mathbf{w}|C)/P(\mathbf{w}|\bar{C})$ by direct observation of $\mathbf{w}$'s in the training set is futile. By making certain independence assumptions [21], we can make the following decomposition:

$$\frac{P(C|\mathbf{w})}{P(\bar{C}|\mathbf{w})} = \frac{P(C)}{P(\bar{C})} \times \prod_{i=1}^{d} \frac{P(w_i|C)}{P(w_i|\bar{C})} \tag{3}$$

Then, using the fact that $P(\bar{C}|\mathbf{w}) = 1 - P(C|\mathbf{w})$, plus some arithmetic manipulations, we can get the following expression for $P(C|\mathbf{w})$:

$$P(C|\mathbf{w}) = \frac{\exp(\log \frac{P(C)}{1-P(C)} + \sum_{i=1}^{d} \log \frac{P(w_i|C)}{P(w_i|\bar{C})})}{1 + \exp(\log \frac{P(C)}{1-P(C)} + \sum_{i=1}^{d} \log \frac{P(w_i|C)}{P(w_i|\bar{C})})} \tag{4}$$

Equation 4 is rarely used directly in text classification, probably because its estimates of $P(C|\mathbf{w})$ are systematically inaccurate. One reason for this inaccuracy is that the independence assumptions made in producing Equation 3 are always incorrect when the $w_i$'s are words or other features defined from natural language. Another problem is that $P(C)$ is typically small and thus hard to estimate, a problem which is compounded when the training set is not a random sample.

Logistic regression [22] provides a partial solution to these problems. It is a general technique for combining multiple predictor values to estimate a posterior probability. The form of the estimate is:

$$P(C|\mathbf{x}) = \frac{\exp(a + b_1 x_1 + ... + b_m x_m)}{1 + \exp(a + b_1 x_1 + ... + b_m x_m)} \tag{5}$$

A number of approaches to using logistic regression in text classification have been proposed [23, 24, 25]. The similarity between Equation 4 and Equation 5 prompted us to try a particularly simple approach, where the log likelihood ratio from the Bayesian independence formulation is used as the single predictor variable:

$$P(C|\mathbf{w}) = \frac{\exp(a + b \sum_{i=1}^{d} \log \frac{P(w_i|C)}{P(w_i|\bar{C})})}{1 + \exp(a + b \sum_{i=1}^{d} \log \frac{P(w_i|C)}{P(w_i|\bar{C})})} \tag{6}$$

Intuitively, we could hope that the logistic parameter $a$ would substitute for the hard-to-estimate prior log odds in Equation 4, while $b$ would serve to dampen extreme log likelihood ratios resulting from independence violations. We have in fact found this simple formulation to work well for text categorization, though we have not compared it with the more complex formulations suggested by other authors. Note that our approach would probably not be appropriate if documents were of widely varying lengths.

---

[3] We are careful to distinguish an example $e$ from the corresponding pattern $\mathbf{w}$, since different examples may have the same feature values $w_1, ..., w_d$.

## 4.2 Training the Classifier

The first step in using Equation 6 is estimating the values $P(w_i|C)/P(w_i|\bar{C})$. We used the following estimator:

$$\frac{P(w_i|C)}{P(w_i|\bar{C})} \doteq \frac{\frac{c_{pi}+(N_p+0.5)/(N_p+N_n+1)}{N_p+d(N_p+0.5)/(N_p+N_n+1)}}{\frac{c_{ni}+(N_n+0.5)/(N_p+N_n+1)}{N_n+d(N_n+0.5)/(N_p+N_n+1)}} \quad (7)$$

Here $N_p$ and $N_n$ are the numbers of tokens in the positive and negative training sets, respectively, $c_{pi}$ and $c_{ni}$ are correspondingly the number of instances of $w_i$ in the positive and negative training sets, and $d$ is the number of features. This is an ad hoc estimator, loosely justified by its analogy to the expected likelihood estimator [26]. The above estimator attempts to avoid extreme estimates of the log likelihood ratio when $N_p$ and $N_n$ are of very different sizes, for instance before our sampling procedure starts finding positive examples.

In text classification there typically is a huge set of potential $w_i$'s, for instance all the types (distinct words) in a collection of documents. Using feature selection to reduce this set (or, equivalently, to lock all but a few values at 0) can improve effectiveness [19]. As a feature quality measure we used:

$$(c_{pi} + c_{ni}) \log \frac{P(w_i|C)}{P(w_i|\bar{C})}. \quad (8)$$

We selected features in order of this value until a specified fraction (0.7 in the experiments reported here) of the total score of all training examples was reached. This was done separately for features with positive and negative log likelihood ratios.

After feature selection is performed, the log likelihood values are used to compute:

$$\sum_{i=1}^{d} \log \frac{P(w_i|C)}{P(w_i|\bar{C})} \quad (9)$$

for each training example. Logistic regression is then used to find the values $a$ and $b$ which give the best fit of this value to the probability of class membership.

## 4.3 Uncertainty Sampling with the Probabilistic Classifier

Uncertainty sampling is simple given a classifier that estimates $P(C|\mathbf{w})$. On each iteration, the current version of classifier can be applied to each example, and those examples with estimated $P(C|\mathbf{w})$ values closest to 0.5 selected, since 0.5 corresponds to the classifier being most uncertain of the class label.

We adopted the slightly more complex method of scoring all examples, and then choosing the $b/2$ examples closest to 0.5 and above it, and the $b/2$ examples closest to 0.5 and below it. This method guarantees that no more than half the examples selected on an iteration are exact duplicates (unless all examples score above 0.5 or all score below 0.5). In addition, there is also some evidence that training on pairs of examples on opposite sides of a decision boundary is useful [14].

## 4.4 Classification with the Probabilistic Classifier

An advantage of using a classifier which provides accurate estimates of $P(C|\mathbf{w})$ is that, under certain assumptions, decision theory gives an optimal rule for deciding whether an example should be assigned to class $C$ ([27], p. 15). Let $l_{ij}$ be the penalty or *loss* incurred for deciding class $i$ when the true class is $j$. (We let $i, j = 1$ for $C$, $i, j = 2$ for $\bar{C}$.) We then should assign the example to class $C$ exactly when:

$$l_{21}P(C|\mathbf{w}) + l_{22}(1 - P(C|\mathbf{w})) > l_{11}P(C|\mathbf{w}) + l_{12}(1 - P(C|\mathbf{w})) \quad (10)$$

For instance, if we desire minimum error rate (both types of incorrect decisions are equally bad) the appropriate losses would be $l_{12} = l_{21} = 1$ and $l_{11} = l_{22} = 0$.

# 5 Experiment

We conducted an experiment to see if uncertainty sampling would reduce the amount of labeled data necessary to train a classifier, in comparison with random sampling and relevance sampling. The training method and probabilistic classifier of Section 4 were used. Classifiers were trained to perform a text categorization task on news story titles.

|            | Training |        | Test   |        |
|------------|----------|--------|--------|--------|
| Category   | Number   | Freq.  | Number | Freq.  |
| tickertalk | 208      | 0.0007 | 40     | 0.0008 |
| boxoffice  | 314      | 0.0010 | 42     | 0.0008 |
| bonds      | 470      | 0.0015 | 60     | 0.0012 |
| nielsens   | 511      | 0.0016 | 87     | 0.0017 |
| burma      | 510      | 0.0016 | 93     | 0.0018 |
| dukakis    | 642      | 0.0020 | 107    | 0.0021 |
| ireland    | 780      | 0.0024 | 117    | 0.0023 |
| quayle     | 786      | 0.0025 | 133    | 0.0026 |
| budget     | 1176     | 0.0037 | 197    | 0.0038 |
| hostages   | 1560     | 0.0049 | 228    | 0.0044 |

**Table 1.** The 10 categories used in our experiments, with number of occurrences and frequency of occurrence on training and test sets.

### 5.1 Data Set

The titles of 371,454 items which appeared on the AP newswire between 1988 and early 1993 were divided randomly into a training set of 319,463 titles and a test set of 51,991 titles. Titles were processed by lower casing text and removing punctuation. Word boundaries were defined by whitespace. Titles were used, rather than the full text of the items, to minimize computation.

Categories to be assigned were based on the "keyword" from the "keyword slug line" present in each AP item ([28], p. 317). The keyword is a string of up to 21 characters indicating the content of the item. While keywords are only required to be identical for updated items on the same news story, in practice there is considerable reuse of keywords and parts of keywords from story to story and year to year, so they have some aspects of a controlled vocabulary.

We defined categories of AP titles according to whether particular substrings appeared in the keyword field. For instance, the following stories were assigned to the *bond* category (the keyword is shown in bold):

**SavingsBonds** Savings Bond Sale Plunge After Rate Cut
**SavingsBonds** Treasury Announces 2 Percent Reduction in Savings Rate
**SavingsBonds-Flood** Flood Victims Permitted to Cash in Savings Bonds Early
**MesaBonds** Mesa to Begin $600 Million Bond Exchange Offer Wednesday
**BondFirms** Report: Wall Street Bond Firms To Ban Political Contributions
**Obit-Bond** James Bond, Ornithologist, Gave Name To Fictional Agent 007
**People-Bond** Julian Bond: Rights Movement Needs Individuals, Not Charismatic Leaders

while these were not:

**Clinton** President-Elect Plays Touch Football
**Bank-Failures** Bank, S&L Failures at Seven-Year Low
**Taxes:SavingsBon** Taxes: Savings Bonds
**TreasuryBorrowing** Treasury Shifts Borrowing Away From Long-Term Bonds
**MuniProbe** Tougher Political Contribution Rules for Muni Bonds Proposed

The categories defined in this fashion were somewhat messy semantically. Julian Bond and James Bond should not be included with savings bonds, while we lose items about financial bonds if the keyword was truncated, misspelled, or emphasized some other aspect of the story. Perfect categorization with these category definitions was therefore not possible. We do not believe this is a serious problem, since we are interested here in the relative rather than absolute effectiveness of categorization methods. Indeed, these categories provide a useful test of the robustness of our methods to errors in the training data.

The 10 categories we defined are shown in Table 1 along with their frequencies in the training and test sets. The categories were chosen to have relatively low frequencies, while still providing a reasonable number of positive examples in both the training and test sets.

### 5.2 Training

The initial classifier required by the uncertainty sampling algorithm (Figure 3) could be produced from a set of words suggested by a teacher, just as classifiers are constructed from the texts of user requests in text retrieval systems [29]. To avoid experimenter bias, we instead used a starting subsample of 3 positive examples of the category randomly selected from the training set. Feature selection always used the words from these 3 examples in addition to words chosen as described in Section 4.2.

On each run, the 3 starting examples were used to train an initial classifier, after which 249 iterations of uncertainty sampling with a subsample size of 4 were carried out as described in Section 4.3. After each subsample was selected, its category labels were looked up and the examples were added to the set of labeled examples to be used in training the next classifier. The classifier produced on each iteration was used for example selection on the next iteration, as well as being retained for evaluation. In order to study the impact of the starting subsample on the quality of the final classifier, we repeated this process 10 times for each category, each time with a different starting subsample of 3 positive examples.

We compared uncertainty sampling with both relevance sampling and random sampling. Relevance sampling was carried out identically to uncertainty sampling, except that the 4 examples with the highest values of $P(C|\mathbf{w})$ were chosen, rather than 4 with values close to 0.5.

The "random" samples actually combined the starting subsamples of 3 positive examples with truly random samples of various sizes. In this fashion, training sets of the following sizes were produced:

3 6 10 20 40 80 160 320 640 1000 2500 4000 6000 8000 10000 15000 20000 30000 40000 50000 60000 70000 80000 ... (by 20000's) ... 300000 319463

Larger sets included the smaller ones. A classifier was formed from each of these sets using the same training methods used for uncertainty and relevance sampling, but the classifier was used only for evaluation, not to guide sampling. Two runs were done from each of 10 starting subsamples for a category, giving a total of 20 runs per category.

### 5.3 Evaluation

We treated each of the 10 categories as a binary classification task and evaluated the classifiers for each category separately. Classifiers were evaluated by applying them with the minimum error rate loss parameters ($l_{12} = l_{21} = 1$ and $l_{11} = l_{22} = 0$) to the 51,991 test items and comparing the classifier decisions with the actual category labels. All classifiers trained on the random samples were evaluated. Classifiers formed during the first 10 iterations of uncertainty and relevance sampling, plus every 5th iteration thereafter, were evaluated.

For text categorization, the effectiveness measures of recall and precision are defined as follows:

$$\text{recall} = \frac{\text{Number of test set category members assigned to category}}{\text{Number of category members in test set}} \quad (11)$$

$$\text{precision} = \frac{\text{Number of test set category members assigned to category}}{\text{Total number of test set members assigned to category}} \quad (12)$$

When comparing two classifiers it is desirable to have a single measure of effectiveness. Van Rijsbergen defined the E-measure as a combination of recall (R) and precision (P) satisfying certain measurement theoretic properties ([30], pp. 168-176):

$$E = 1 - \frac{(\beta^2 + 1)PR}{\beta^2 P + R} \quad (13)$$

The parameter $\beta$ ranges between 0 and infinity and controls the relative weight given to recall and precision. A $\beta$ of 1 corresponds to equal weighting of recall and precision. To get a single measure of effectiveness where higher values correspond to better effectiveness, and where recall and precision are of equal importance, we define $F_{\beta=1} = 1 - E_{\beta=1}$.

## 6 Results

Table 2 shows for each category the mean $F_{\beta=1}$ for classifiers formed by uncertainty sampling as well as those formed by relevance sampling and on the full training set. We also show effectiveness using just the 3 starting examples, plus 7 randomly selected examples. This gives a sense of the quality of the initial classifier. For all categories except *tickertalk*, an uncertainty sample of 999 texts resulted in a classifier substantially more effective than the initial classifier or one formed from a relevance sample of 999 texts. The classifier was usually of similar or better effectiveness than one trained on all 319,463 texts. Classifiers trained on a random sample of 1000 texts in most cases had very low effectiveness.

Figure 2 plots effectiveness against sample size for uncertainty sampling, relevance sampling, and random sampling. Results for 9 categories are presented, omitting *tickertalk* on which no strategy worked well.

|  | 3 + 996 uncer. | | 3 + 7 rand. | | 3 + 996 rel. | | 3 + 319,460 full | |
|---|---|---|---|---|---|---|---|---|
| Category | mean | SD | mean | SD | mean | SD | mean | SD |
| tickertalk | .033 | (.031) | .018 | (.023) | .023 | (.039) | .047 | (.001) |
| boxoffice | .700 | (.041) | .222 | (.172) | .481 | (.053) | .647 | (.023) |
| bonds | .636 | (.034) | .146 | (.134) | .541 | (.069) | .509 | (.020) |
| nielsens | .801 | (.016) | .291 | (.218) | .567 | (.132) | .741 | (.022) |
| burma | .653 | (.035) | .032 | (.033) | .201 | (.057) | .464 | (.023) |
| dukakis | .136 | (.046) | .101 | (.075) | .035 | (.021) | .163 | (.015) |
| ireland | .416 | (.041) | .050 | (.033) | .170 | (.038) | .288 | (.030) |
| quayle | .386 | (.040) | .081 | (.064) | .140 | (.072) | .493 | (.009) |
| budget | .290 | (.039) | .058 | (.046) | .141 | (.029) | .235 | (.005) |
| hostages | .477 | (.021) | .068 | (.042) | .177 | (.039) | .498 | (.003) |

**Table 2.** Mean and standard deviation of $F_{\beta=1}$ for training on initial 3 examples combined with each of 996 uncertainty selected examples, 7 random examples, 996 relevance selected examples, or 319,460 remaining examples. Means are over 10 runs for uncertainty and relevance sampling, and over 20 runs for random and full sampling.

## 7 Discussion

As Figure 2 shows, classifier effectiveness generally increases with sample size under all sampling methods, but faster with the two sequential methods. Of the sequential methods, uncertainty sampling substantially outperforms relevance sampling. The results hold across categories with widely varying absolute levels of effectiveness.[4]

The superiority of uncertainty sampling over relevance sampling is particularly notable since the low frequency of the categories used limited the danger that relevance sampling would drown in positive examples. Indeed, the difference between uncertainty sampling and relevance sampling is lower for the less frequent categories. However, even here uncertainty sampling is better, both in its higher mean and in its lower standard deviation.

In most cases effectiveness levels reached by random sampling only with 100,000 or more training examples are reached by uncertainty sampling with less than 1000 examples, while training on 1000 randomly selected examples gives greatly inferior results. In some cases, reaching a given level of effectiveness requires 500 times as many randomly selected examples as examples selected by uncertainty sampling.

Some caution is required in comparing the results for small uncertainty samples with those for large random samples. For 6 of 10 categories, the mean $F_{\beta=1}$ for a classifier trained on a uncertainty sample of 999 examples actually exceeds that from training on the full training set of 319,463 examples. This means that some aspect of our classifier training is not making effective use of large training sets. Feature selection is the most likely villain. Our method produced several thousand features when applied to the full training set, and previous work suggests this is too many [19].

The graphs of average effectiveness hide some variation from run to run. Several of the standard deviations shown in Table 2 amount to 10% or more of the mean effectiveness, meaning that the quality of the final classifiers is somewhat unpredictable.

One source of this variation was our initial subsamples of 3 positive examples. When a subsample was badly unrepresentative of the category, the initial classifier was ineffective, and there was a considerable delay before uncertainty sampling started to find additional positive examples. Even initial classifiers with the same raw effectiveness could lead to different parts of the space of examples being searched first.

Besides influencing early classifier formation, the starting subsamples had an impact on effectiveness through our making their words required features. This can be seen in the standard deviations for the Full column of Table 2 where, since all classifiers were trained on the same set of examples, the only difference among runs for a category is in the required features provided by the initial examples.

A second source of variation was fluctuations in the quality of successive classifiers. The uncertainty sampling process is inherently exploratory, with deficiencies in the classifier produced on one iteration leading to the selection of compensating examples on later iterations.

## 8 Future Work

Our results demonstrate that effective text classifiers can be created by obtaining large amounts of unlabeled data and labeling only a small fraction of it. While we tested uncertainty sampling on a text categorization task, it can equally well be applied to any classification task.

---

[4] The variations in absolute levels of effectiveness are to be expected—some categories are simply harder than others, and some of our category definitions were particularly noisy.

Text retrieval is an obvious application, though the tradeoff between retrieving the texts most useful to the user vs. the texts from which the system will learn the most must be considered [31]. The tradeoff is less of an issue in filtering, routing, and information dissemination applications, since the cost of judging nonrelevant examples can be amortized over a longer period of operation.

Uncertainty sampling should also benefit classification-based approaches to natural language processing tasks. A number of projects have annotated or are annotating large corpora to support the training of statistical methods for these tasks. Our results suggest that gathering huge unlabeled corpora, and using uncertainty sampling to annotate a small subset for each task may be cheaper and equally effective. Domains besides text processing where large data sets are available can also benefit.

Many questions remain to be answered about uncertainty sampling. The most important practical ones have to do with how the teacher knows when to stop, and how to form the final classifier for use after they do. Estimates of classifier effectiveness would enable the teacher to track progress, but new methods will be needed to produce such an estimate from a sample which is not random. Effectiveness estimates would also aid in selecting a classifier to use from the classifiers formed on the last few iterations. Alternately, methods for stabilizing the fluctuations from iteration to iteration might be pursued.

A variety of extensions and improvements to uncertainty sampling can be explored. We need to determine the relationship between subsample size and effectiveness, since larger subsamples require less computation. It also seems likely that subsample size can be increased if redundancy within subsamples is decreased. Other efficiency improvements include using a less accurate but more efficiently trained classifier during sampling [32], and picking the first examples satisfying a threshold on uncertainty rather than the most uncertain examples. Simultaneous training of classifiers for multiple classes is also of interest.

As currently formulated, uncertainty sampling requires that the underlying training algorithm produce reasonable classifiers even from very small training sets. This meant we had to tinker with feature selection and parameter estimation to avoid producing pathological behavior on small samples. We are currently exploring variations on uncertainty sampling which would be more robust with respect to problems in classifier training.

## 9 Summary

Text is cheap, but information, in the form of knowing what classes a text belongs to, is expensive. Automatic classification of text can provide this information at low cost, but the classifiers themselves must be built with expensive human effort, or trained from texts which have themselves been manually classified. We have demonstrated that uncertainty sampling can sharply reduce the amount of text which must be manually labeled to create an effective classifier. Uncertainty sampling has potential applications in a variety of text processing tasks, as well as in other domains where large amounts of unclassified data are available.

## 10 Acknowledgments

We thank Jason Catlett, William Cohen, Eileen Fitzpatrick, Yoav Freund, Trevor Hastie, Robert Schapire, and Sebastian Seung for advice and useful comments on this work, and Ken Church for making available his text processing tools and his help with them.

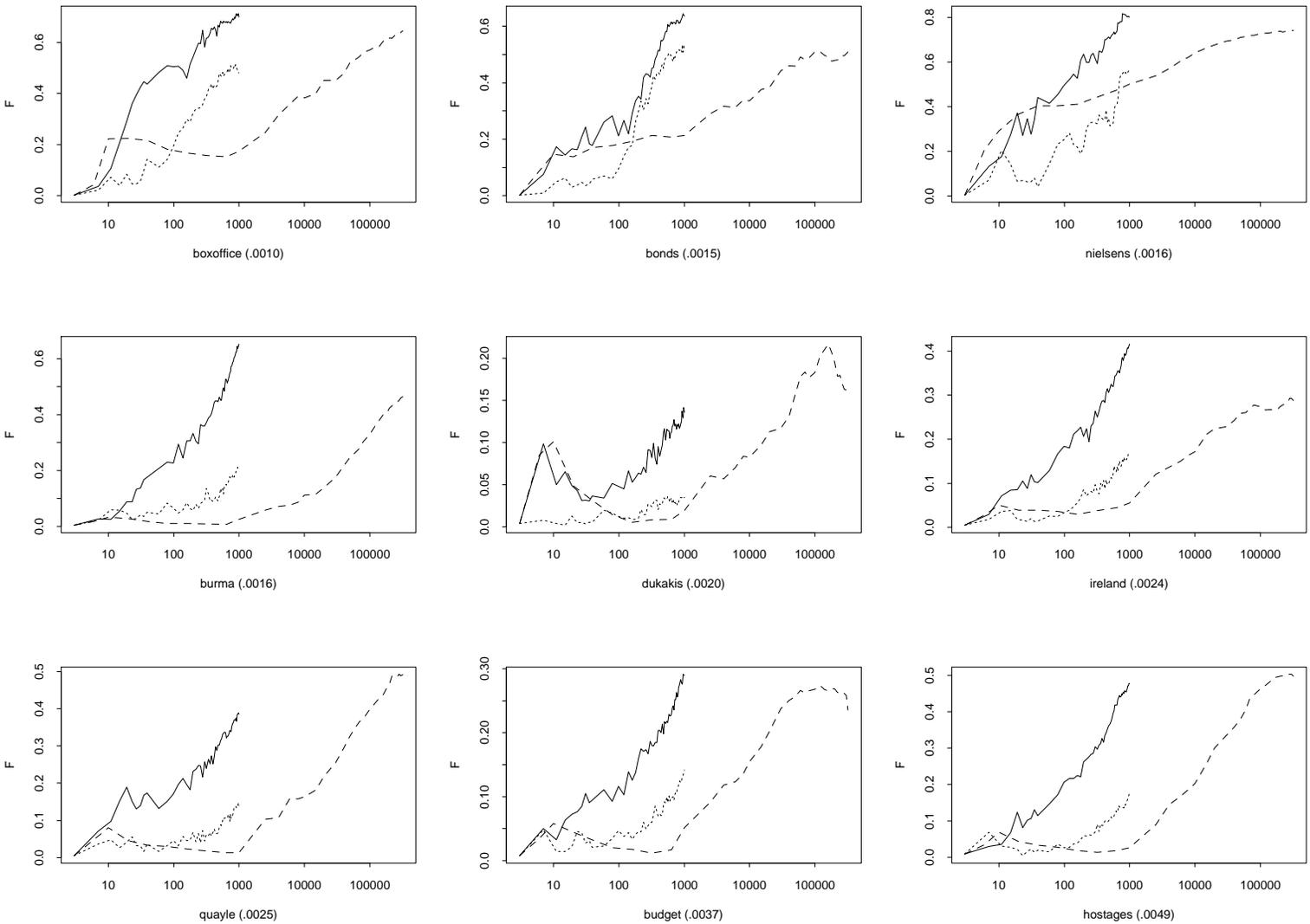

**Figure 2**. Mean $F_{\beta=1}$ values for text classifiers trained on uncertainty (solid line), relevance (dotted line) and random (dashed line) samples of titles from AP corpus. Means are over 10 runs for uncertainty and relevance sampling, and over 20 runs for random sampling. Results shown for 9 categories. Frequency of category on training set shown in parentheses.